\documentclass[10pt]{article}
\usepackage[T1]{fontenc}
\usepackage[margin=1in]{geometry}
\usepackage[numbers,sort&compress]{natbib}
\usepackage{authblk}
\usepackage{microtype}
\usepackage{graphicx}
\usepackage{subcaption}
\usepackage{booktabs}
\usepackage{array}
\usepackage[table]{xcolor}
\usepackage{adjustbox}
\usepackage{amsmath,amssymb,mathtools}
\usepackage[capitalize,noabbrev]{cleveref}
\usepackage{url}
\usepackage{algorithm}
\usepackage{algpseudocode}
\usepackage{enumitem}
\usepackage{caption}
\usepackage{placeins}
\usepackage{float}

\providecommand{\keywords}[1]{\par\noindent\textbf{Keywords:} #1}

\graphicspath{{figures/}}
\newcommand{\method}{\textsc{LifelongVLA}}
\newcommand{\best}[1]{\textcolor{red}{\textbf{#1}}}
\newcommand{\second}[1]{\textcolor{blue}{\textbf{#1}}}

\setlist[itemize]{leftmargin=1.1em,itemsep=0.1em,topsep=0.2em}
\setlist[enumerate]{leftmargin=1.2em,itemsep=0.1em,topsep=0.2em}

\title{Towards Human-like Physical Intelligence: Lifelong Vision-Language-Action Learning for Robotic Manipulation}
\author[1]{Yao He}
\author[1]{Gan Sun\thanks{Corresponding author.}}
\author[2]{Wenqi Liang}
\author[1]{Fazeng Li}
\author[1]{Yang Cong}
\affil[1]{South China University of Technology}
\affil[2]{University of Trento}

\date{}
\begin{document}
\maketitle

\begin{abstract}
Similar to the natural capabilities of humans to sequentially learn new tasks, robots with Vision-Language-Action (VLA) model should possess lifelong learning ability to learn a new task when deployed for open-world environment. However, most recently-proposed lifelong learning models aim to effectively learn the current task (plasticity) or maintain high accuracy on the previous tasks (stability), the plasticity-stability trade-off remains largely unsolved in robot manipulation models. To address this fundamental challenge, we propose a cache-efficient lifelong Vision-Language-Action learning framework for robotic manipulation (\emph{i.e.,} \method{}), which alleviates plasticity-stability trade-off with a dual-timescale adaptation mechanism while achieving low-cost robotic deployment with a cache-efficient replay strategy. More concretely, we propose a dual-timescale LoRA gating module to decompose VLA adaptation into two lightweight pathways: a short-term adapter for plasticity and a long-term adapter for stable consolidation. These pathways are integrated via a task-aware gate, which could enable explicit control of the plasticity--stability trade-off. In the skill replay phase, a cache-efficient stochastic replay strategy is proposed to preserve more balanced retention signals without full-trajectory storage. To the end, experiments show that \method{} outperforms existing baselines, demonstrating efficient skill expansion, robust retention of learned manipulation behaviors, and less reliance on retraining reliance for real-world deployment on xArm robot.

\end{abstract}

\keywords{Lifelong Learning, Vision-Language-Action Models, Robot Manipulation}

\section{Introduction}
\label{sec:introduction}
Vision-Language-Action (VLA) policies~\citep{kim2025srt,lin2025showui,wen2025tinyvla,yao2025long} have emerged as a key paradigm for language-conditioned robotic manipulation by connecting pretrained vision--language representations with robot action generation. Given visual observations and natural language instructions, VLA policies produce actions for embodied control, extending multimodal understanding from passive perception to physical interaction. Recent systems, including RT-style models~\citep{brohan2023rt2}, Octo~\citep{octo2024}, $\pi_0$~\citep{pi2024pi0}, $\pi_{0.5}$~\citep{pi2025pi05}, and $\pi_{0.7}$~\citep{physicalintelligence2026pi07}, further scale this paradigm with large robot datasets and vision--language pretraining. These advances bridge high-level semantic reasoning and low-level motor control through multimodal alignment and policy learning.

Although these advances have improved language-conditioned manipulation, most VLA policies still follow a static offline training paradigm: they are trained once on a fixed set of pre-collected skills and deployed without further updates~\citep{kim2024openvla,liang2025discrete,liang2025pixelvla}. Similar to the natural ability of humans to learn new tasks sequentially, robots should be able to acquire new manipulation skills continually after deployment. However, this paradigm is poorly suited to real-world robot deployment, where new objects, instructions, scene layouts, and embodiment-specific action distributions emerge continuously~\citep{li2026remem,zheng2025imanip,liang2024never}. Lifelong learning or continual learning is therefore essential for enabling VLA policies to acquire new capabilities while preserving previously learned ones. A straightforward solution is to learn new skills sequentially, but direct sequential learning can cause catastrophic forgetting of previously acquired knowledge and behaviors~\citep{rebuffi2024incremental,wang2024comprehensive,shenfeld2025rl,lai2025reinforcement}. Alternatively, retaining all previous task data and repeatedly retraining over the full history can mitigate forgetting. It incurs prohibitive memory and computational costs as the number of tasks grows.

\begin{figure}[t]
  \centering
  \includegraphics[width=0.96\linewidth]{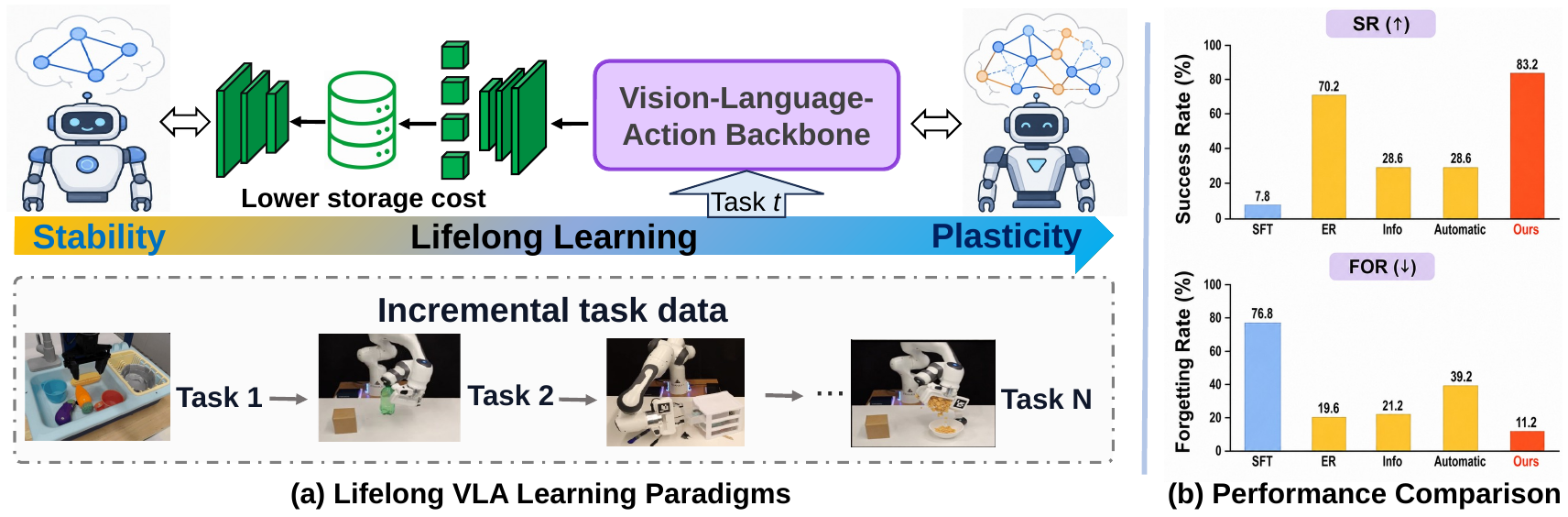}
  \vspace{-0.2em} 
\caption{\textbf{Motivation and overview.} (a) Lifelong VLA learning with incremental tasks, stability--plasticity balance, and low-storage replay. (b) Performance comparison of \method{} with competing baselines.}
  \label{fig:intro_overview}
  \vspace{-10pt}
\end{figure}

To adapt into such real-world deployment scenarios, existing skill-incremental methods often rely on replay buffers, architectural expansion, or parameter isolation~\citep{li2026remem,zheng2025imanip,liang2024never}. However, these methods still leave the plasticity-stability dilemma insufficiently addressed~\citep{dohare2024loss,zhao2025mllm,sokar2025continual}. High plasticity enables rapid adaptation to new skills but may overwrite learned behaviors, whereas strong stability preserves past skills but can limit future learning. Recent studies further show that simple Experience Replay works surprisingly well for VLA policies by revisiting past experience during continual updates~\citep{liu2026pretrained}, while these solutions remain costly for large VLA backbones. Full-trajectory replay requires storing and loading image-rich episodes, which introduces expensive backbone computation via repeated re-encoding of old observations. These expansion-based designs can increase routing cost and inference latency. To efficiently acquire new tasks while preserving prior skills, lifelong learning for large VLA models requires both an explicit mechanism for balancing plasticity and stability and a cache-efficient design that reduces memory and computation overhead.

To address such challenging scenarios, we take an earlier attempt to explore lifelong vision-language-action (VLA) learning, which enables robots to continually acquire new manipulation skills while preserving their capabilities on previous tasks. Unlike the conventional continual learning paradigm, this lifelong VLA learning centers on a fundamental plasticity-stability trade-off: how to acquire incoming manipulation skills without eroding previously learned behaviors, as illustrated in Fig.~\ref{fig:intro_overview}(a). We then propose \method{}, a cache-efficient lifelong vision-language-action learning method built on a frozen backbone to balance plasticity and stability under limited replay. Its first key module, \textbf{Dual-Timescale LoRA Gating}, separates adaptation into two LoRA pathways with different time scales: a short-term pathway for rapidly learning newly arriving skills and a long-term pathway for consolidating previously learned skills over time. A task-aware gate composes the two pathways at the weight level, allowing the policy to dynamically adjust the contribution of short-term plastic updates and long-term stable updates during adaptation. This design provides an explicit interface for balancing plasticity and stability without modifying the backbone or requiring two forward passes. To reduce rehearsal overhead, \method{} introduces a \textbf{Cache-Efficient Stochastic Replay} module to keep only a few randomly selected samples from each skill as stop-gradient prefix tokens, together with compact state--action supervision. During replay, old samples are resampled into multiple stochastic denoising instances under the current model state, allowing a small cache to provide diverse and balanced retention signals. This design turns replay from expensive full-history rehearsal into compact memory-guided adaptation, enabling efficient skill expansion while preserving old behaviors. As shown in Fig.~\ref{fig:intro_overview}(b), \method{} improves new-skill acquisition and reduces forgetting compared with existing baselines, with over 13\% improvement in success rate and over 8.2\% reduction in forgetting rate, while reducing the replay memory cost by 91.92 MiB per task.


The contributions of this work are listed below: i) we introduce \method{}, a lifelong-learning VLA method inspired by the human ability to learn tasks sequentially, enabling open-world robots to continually acquire new knowledge and skills, balance new-skill plasticity with old-skill stability and reduce replay overhead with a cache-efficient design; ii) we propose a \textbf{Dual-Timescale LoRA Gating} module, which uses short-term and long-term LoRA pathways with a task-aware weight-level gate to explicitly control the plasticity-stability trade-off in a single forward pass; and iii) we develop a \textbf{Cache-Efficient Stochastic Replay} module, which stores only a few randomly selected samples from each skill and resamples them into stochastic denoising instances, providing rich retention signals without full-trajectory storage or large-scale replay.

\section{Related Work}
\label{sec:related_work}

\paragraph{Vision--Language--Action models.}
VLA models ground language instructions in visual observations and generate executable robot actions. Recent progress has been driven by scaling transformer policies with large multi-task robot datasets and transferring web-scale vision--language knowledge into embodied control. RT-style models scale real-robot behavior learning, while RT-2 and PaLM-E inject semantic representations from VLMs and LLMs into robotic decision making~\citep{brohan2022rt1,openx2023,brohan2023rt2,driess2023palme}. Open-source and cross-embodiment efforts broaden hardware coverage and improve generalization across tasks and robots~\citep{octo2024,bousmalis2023robocat,reed2022gato}. Other methods combine language reasoning with feasibility-aware execution or planning, including modular pipelines and end-to-end visuomotor policies~\citep{shridhar2021cliport,shridhar2022peract,jiang2023vima,ahn2022saycan,liang2023cap,huang2023voxposer}. More recent action-generation approaches move beyond discrete token prediction toward continuous generative policies and stronger action representations~\citep{chi2023diffusionpolicy,liu2024dp4,pi2024pi0,pi2025pi05,pertsch2025fast,guruprasad2025benchmarkvla,shukor2025smolvla}. Despite this progress, most VLA systems assume a relatively fixed skill repertoire. Continued training on new skills can introduce cross-skill interference and regression, leaving open-ended lifelong acquisition unresolved.

\paragraph{Continual learning.}
Continual learning aims to incorporate new tasks over time while preserving performance on previous ones. Classical approaches use importance-aware regularization, distillation, rehearsal, architectural expansion, or parameter isolation to reduce interference~\citep{kirkpatrick2017ewc,li2017lwf,rusu2016progressive,mallya2018packnet}. With the rise of frozen foundation models, rehearsal-free and parameter-efficient strategies have become increasingly relevant, including prompt-based methods, modular experts, and adapter routing~\citep{wang2022dualprompt,smith2023codaprompt,yu2024moeadapter}. These methods align naturally with PEFT modules such as LoRA, prefix tuning, and prompt tuning~\citep{hu2022lora,li2021prefix,lester2021prompttuning}. However, generic continual learning methods are not directly tailored to VLA policies, where long-horizon control, temporally coupled actions, large multimodal backbones, and expensive trajectory replay create distinct constraints~\citep{wolczyk2021continualworld}. \method{} focuses on this setting by combining two-time-scale adapter learning, continuous adapter composition, and lightweight feature-level retention.

\section{Preliminary and Problem Definition}
\label{sec:problem}

\paragraph{Vision--Language--Action policy.}
A Vision--Language--Action (VLA) policy~\citep{cui2025openhelix,niu2024llarva} maps a language instruction $p$, visual observations $x=\{x^k\}_{k=1}^{K}$, and robot states $s=\{s^k\}_{k=1}^{K}$ to a continuous action sequence $y=\{a^k\}_{k=1}^{K}$. We denote the full VLA input as $u=(x,p,s)$. Let $F_{\Theta_0}$ be a pretrained VLA backbone with frozen parameters $\Theta_0$. We view the VLA policy as a prefix encoder $E_{\mathrm{pre}}$ for visual-language context, a suffix encoder $E_{\mathrm{suf}}$ for state and noised action inputs, and an action decoder $D$ for predicting the denoising target. In our implementation, the continuous action decoder follows a diffusion-style action prediction objective, where noised actions are generated using sampled diffusion variables. To adapt the backbone efficiently, we apply LoRA to the adapted linear layers. For the $\ell$-th layer, the frozen weight $W_0^\ell$ is updated by a low-rank residual, $W_t^\ell=W_0^\ell+\gamma\Delta W_t^\ell$, where $\Delta W_t^\ell=A_t^\ell B_t^\ell$. Here, $A_t^\ell$ and $B_t^\ell$ are learnable low-rank factors with rank $r$, and $\gamma$ is the shared LoRA scaling factor. The adapted parameter set at task $t$ is written as $\Theta_t=\Theta_0+\Delta\Theta_t$, where $\Delta\Theta_t=\{\Delta W_t^\ell\}_{\ell=1}^{L}$.

\paragraph{LifelongVLA setting.}
We consider an online sequence of manipulation tasks~\citep{meng2025preserving} $\mathcal{T}=\{\mathcal{T}_1,\mathcal{T}_2,\ldots,\mathcal{T}_T\}$, where each task corresponds to one manipulation skill. For task $t$, only the current dataset $\mathcal{S}_t=\{(u_i^t,y_i^t)\}_{i=1}^{N_t}$ is available for training. Previous datasets are unavailable except for a bounded cache, future task distributions are unknown, and task identities are not provided during inference. The goal is to learn the current manipulation skill while preserving previously acquired skills under limited memory and computation.

\begin{figure*}[t]
  \centering
  \includegraphics[width=0.96\textwidth]{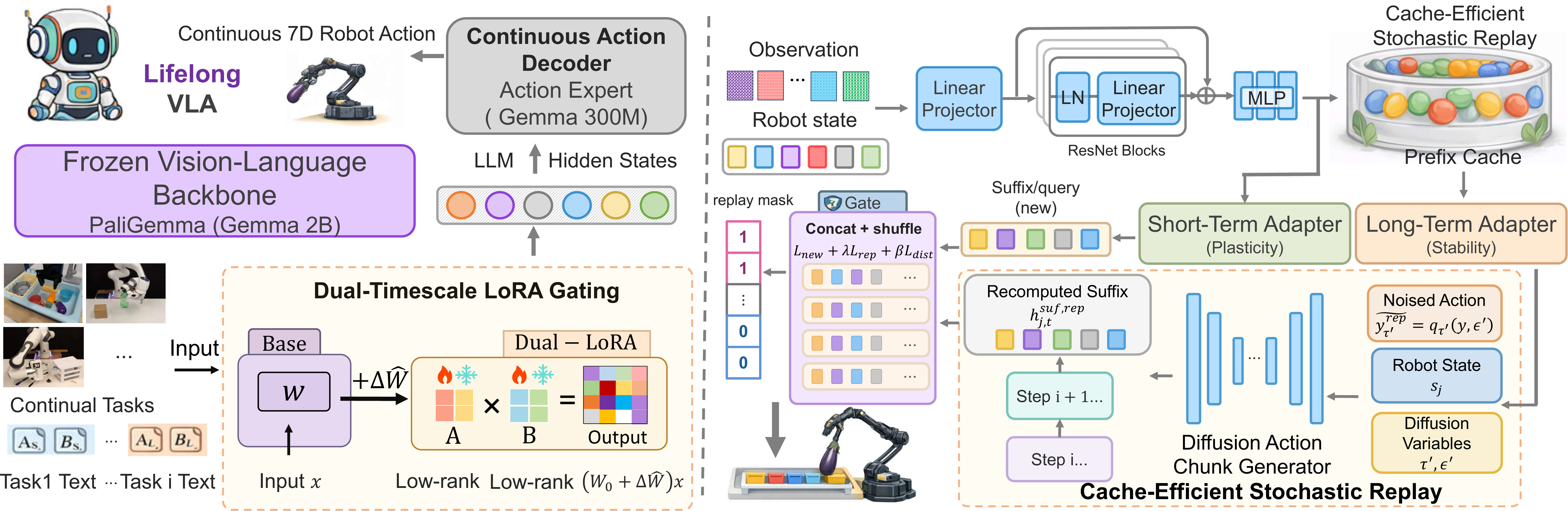}
\caption{\textbf{Algorithmic diagram of \method{}}, which consists of two modules: (i) \textbf{Dual-Timescale LoRA Gating}, which composes short-term and long-term LoRA pathways with a task-aware gate to balance new-skill plasticity and old-skill stability, and (ii) \textbf{Cache-Efficient Stochastic Replay}, which stores compact prefix tokens and reconstructs replay suffixes to provide old-skill supervision without full-trajectory replay.}
  \label{fig:method_overview}
\end{figure*}

\section{The Proposed Model}
\label{sec:method}

As shown in \cref{fig:method_overview}, the pipeline of \method{} is designed to address the lifelong VLA learning problem. Specifically, we design \textbf{Dual-Timescale LoRA Gating} to separate fast plastic adaptation and stable memory into short- and long-term LoRA pathways, which are composed by a task-aware gate to explicitly balance new-skill learning and old-skill retention. Moreover, we develop \textbf{Cache-Efficient Stochastic Replay} to store only stop-gradient prefix tokens with compact state--action supervision and reconstruct replay suffixes under fresh diffusion variables, thereby reducing replay overhead while preserving old-skill supervision. The details are presented below.

\subsection{Dual-Timescale LoRA Gating}
\label{sec:dual_timescale}

Lifelong VLA learning requires a policy to rapidly absorb new task knowledge while retaining previously learned manipulation behaviors. A single LoRA adapter uses the same update pathway for both goals, which can entangle new-skill plasticity with old-skill stability and make adaptation difficult to control. \method{} addresses this issue by decomposing adaptation into short- and long-term LoRA pathways, so that recent task-specific changes and stable skill memory can be modeled separately.

\paragraph{Short- and long-term pathways.}
For each adapted layer $\ell$, \method{} defines a short-term update and a long-term update as
\begin{equation}
\begin{aligned}
    \Delta W_{\mathrm{sh},t}^{\ell}
    &=
    A_{\mathrm{sh},t}^{\ell}B_{\mathrm{sh},t}^{\ell},\\
    \Delta W_{\mathrm{lg},t}^{\ell}
    &=
    A_{\mathrm{lg},t}^{\ell}B_{\mathrm{lg},t}^{\ell}.
\end{aligned}
\label{eq:short_long_lora}
\end{equation}
The two pathways are optimized with different supervision signals. The short-term LoRA pathway is updated by the current-task loss to rapidly learn newly arriving skills, while the long-term LoRA pathway is updated only with replay and distillation signals to consolidate previously learned skills:
\begin{equation}
    \psi_{\mathrm{sh},t}
    \leftarrow
    \psi_{\mathrm{sh},t}
    -
    \eta_{\mathrm{sh}}
    \nabla_{\psi_{\mathrm{sh}}}
    \mathcal{L}_{\mathrm{new}},
    \qquad
    \psi_{\mathrm{lg},t}
    \leftarrow
    \psi_{\mathrm{lg},t}
    -
    \eta_{\mathrm{lg}}
    \nabla_{\psi_{\mathrm{lg}}}
    \left(
    \lambda\mathcal{L}_{\mathrm{rep}}
    +
    \beta\mathcal{L}_{\mathrm{dist}}
    \right),
    \label{eq:dual_update}
\end{equation}
where $\psi_{\mathrm{sh},t}$ and $\psi_{\mathrm{lg},t}$ denote the parameters of the short- and long-term LoRA pathways, respectively. Here, $\mathcal{L}_{\mathrm{new}}$ denotes the current-task denoising loss, $\mathcal{L}_{\mathrm{rep}}$ denotes the replay loss on cached old-skill samples, $\mathcal{L}_{\mathrm{dist}}$ denotes the teacher distillation loss, and $\beta$ controls the strength of distillation. Together with $\eta_{\mathrm{lg}}<\eta_{\mathrm{sh}}$, this makes the long-term pathway evolve more slowly than the short-term pathway. This design gives the short-term pathway high plasticity for recent task-specific changes, while the long-term pathway maintains stable manipulation knowledge through replay and distillation.

\paragraph{Shared gate.}
If the gate context were computed from features after gated adaptation, the gate would depend on the very weights it controls. To avoid this circular dependency, the gate context $c$ is computed from frozen prefix features before applying the gated LoRA weights:
\begin{equation}
    c
    =
    \operatorname{Pool}
    \left(
    \mathrm{stopgrad}
    \left(
    E_{\mathrm{pre}}(u;\Theta_0)
    \right)
    \right).
    \label{eq:gate_context}
\end{equation}
Given this gate context, the shared gate predicts the contribution of the long-term pathway:
\begin{equation}
    \alpha_t^\ell(c)
    =
    \sigma
    \left(
    W_g^\ell \phi^\ell(c)+b_g^\ell
    \right),
    \label{eq:shared_gate}
\end{equation}
where $\phi^\ell(\cdot)$ is a lightweight feature projection, $W_g^\ell,b_g^\ell$ are gate parameters, and $\alpha_t^\ell(c)\in[0,1]$. The gate parameters are shared for current and replay samples, while the context $c$ is sample-dependent.

\paragraph{Weight-level composition.}
The two LoRA pathways are composed directly in weight space:
\begin{equation}
\begin{split}
    \Delta \widehat{W}_{t}^{\ell}(c)
    &=
    \big(1-\alpha_t^\ell(c)\big)
    \Delta W_{\mathrm{sh},t}^{\ell}
    +
    \alpha_t^\ell(c)
    \Delta W_{\mathrm{lg},t}^{\ell},\\
    \widehat{W}_{t}^{\ell}(c)
    &=
    W_0^\ell
    +
    \gamma
    \Delta \widehat{W}_{t}^{\ell}(c).
\end{split}
\label{eq:gated_lora}
\end{equation}
Thus, $\alpha_t^\ell(c)\rightarrow 0$ emphasizes recent plastic adaptation, whereas $\alpha_t^\ell(c)\rightarrow 1$ emphasizes stable memory. Since the composition is performed on LoRA weights, inference still requires only a single forward pass through the adapted VLA.

We denote the resulting context-conditioned adapted parameters by
\begin{equation}
    \widehat{\Theta}_{t}(c)
    =
    \Theta_0
    +
    \{\gamma\Delta \widehat{W}_{t}^{\ell}(c)\}_{\ell=1}^{L}.
    \label{eq:context_params}
\end{equation}

\subsection{Cache-Efficient Stochastic Replay}
\label{sec:stochastic_replay}

Full trajectory replay is costly for VLA policies~\citep{li2026remem} because it requires storing image-rich episodes and repeatedly re-encoding old observations. Dense feature replay reduces storage, but cached suffix features or fixed diffusion queries can become stale as the adapted model changes. \method{} instead stores compact prefix tokens as old-skill memory anchors and recomputes suffix tokens with the current model during replay.

\paragraph{Skill-level prefix cache.}
Let $\Omega_t$ be all eligible samples from task $\mathcal{T}_t$. For each new task, \method{} randomly selects a small skill-level subset $\mathcal{R}_t\subseteq\Omega_t$ with $|\mathcal{R}_t|=\min(c_t,|\Omega_t|)$, where $c_t$ is the cache quota. For each selected sample, the replay buffer stores only the stop-gradient prefix token and compact state--action supervision:
\begin{equation}
    \mathcal{B}_t
    =
    \operatorname{Reservoir}_{M}
    \left(
    \mathcal{B}_{t-1}
    \cup
    \left\{
    \big(
    \bar{h}_{i,t}^{\mathrm{pre}},
    s_i^t,
    y_i^t
    \big)
    :
    i\in\mathcal{R}_t
    \right\}
    \right),
    \qquad
    \bar{h}_{i,t}^{\mathrm{pre}}
    =
    \mathrm{stopgrad}
    \left(
    h_{i,t}^{\mathrm{pre}}
    \right).
    \label{eq:prefix_cache}
\end{equation}
The cache does not store raw images, language tokens, suffix tokens, diffusion time, or diffusion noise. Compared with full-trajectory replay, this design substantially reduces storage and I/O overhead.

\paragraph{Stochastic replay with recomputed suffix.}
For a replay sample $(\bar{h}_{j}^{\mathrm{pre}},s_j,y_j)\sim\mathcal{B}_{t-1}$, \method{} freshly re-samples diffusion variables and constructs a noised action:
\begin{equation}
    \tau_j'\sim p(\tau),
    \qquad
    \epsilon_j'\sim\mathcal{N}(0,I),
    \qquad
    \tilde{y}_{\tau_j'}^{\mathrm{rep}}
    =
    q_{\tau_j'}(y_j,\epsilon_j').
    \label{eq:rep_noise}
\end{equation}
The cached prefix is then combined with a suffix token recomputed by the current model,
\begin{equation}
    h_{j,t}^{\mathrm{suf,rep}}
    =
    E_{\mathrm{suf}}
    \left(
    s_j,\tilde{y}_{\tau_j'}^{\mathrm{rep}},\tau_j';
    \widehat{\Theta}_{t}(c_j^{\mathrm{rep}})
    \right),
    \qquad
    \hat{\eta}_{j}^{\mathrm{rep}}
    =
    D
    \left(
    [\bar{h}_{j}^{\mathrm{pre}},h_{j,t}^{\mathrm{suf,rep}}];
    \widehat{\Theta}_{t}(c_j^{\mathrm{rep}})
    \right).
    \label{eq:rep_forward}
\end{equation}
Because $\tau_j'$ and $\epsilon_j'$ are re-sampled at every replay step, each cached prefix can generate multiple stochastic replay instances. The prefix serves as an old-skill memory anchor, while the recomputed suffix reduces stale mismatch between cached and current representations.

\paragraph{Masked new--replay objective.}
Current samples and replay samples are concatenated and shuffled in the same mini-batch. Let $\mathcal{M}$ denote the mixed mini-batch, and let $m_i\in\{0,1\}$ be the replay mask, where $m_i=0$ denotes a current sample and $m_i=1$ denotes a replay sample. We define the normalized current-task loss and replay loss as
\begin{equation}
    \mathcal{L}_{\mathrm{new}}
    =
    \frac{1}{|\mathcal{M}_{\mathrm{new}}|}
    \sum_{i\in\mathcal{M}}
    (1-m_i)
    \ell
    \left(
    \hat{\eta}_{i},
    \eta_i^{\star}
    \right),
    \qquad
    \mathcal{L}_{\mathrm{rep}}
    =
    \frac{1}{|\mathcal{M}_{\mathrm{rep}}|}
    \sum_{i\in\mathcal{M}}
    m_i
    \ell
    \left(
    \hat{\eta}_{i},
    \eta_i^{\star}
    \right),
    \label{eq:masked_new_rep_loss}
\end{equation}
where $\mathcal{M}_{\mathrm{new}}=\{i\in\mathcal{M}:m_i=0\}$ and $\mathcal{M}_{\mathrm{rep}}=\{i\in\mathcal{M}:m_i=1\}$.

For replay samples, we further use the detached model snapshot $\bar{F}_{t-1}$ from task $t-1$ as a teacher to stabilize old-skill predictions. The teacher prediction is computed by $\bar{F}_{t-1}$ using the same replay tuple and sampled diffusion variables, with all teacher outputs detached from the computation graph. Let $\hat{\eta}_{i}^{\mathrm{old}}$ be the detached teacher prediction. The distillation loss is
\begin{equation}
    \mathcal{L}_{\mathrm{dist}}
    =
    \frac{1}{|\mathcal{M}_{\mathrm{rep}}|}
    \sum_{i\in\mathcal{M}}
    m_i
    \left\|
    \hat{\eta}_{i}
    -
    \mathrm{stopgrad}
    \left(
    \hat{\eta}_{i}^{\mathrm{old}}
    \right)
    \right\|_2^2.
    \label{eq:dist_loss}
\end{equation}
The final masked objective is
\begin{equation}
    \mathcal{L}_{t}
    =
    \mathcal{L}_{\mathrm{new}}
    +
    \lambda
    \mathcal{L}_{\mathrm{rep}}
    +
    \beta
    \mathcal{L}_{\mathrm{dist}}.
    \label{eq:masked_stage_loss}
\end{equation}
The separate normalization of current and replay terms prevents either source from dominating the objective due to relative batch size, yielding a balanced retention signal under a small replay budget.
\begin{figure*}[t]
  \centering
  \includegraphics[width=0.96\textwidth]{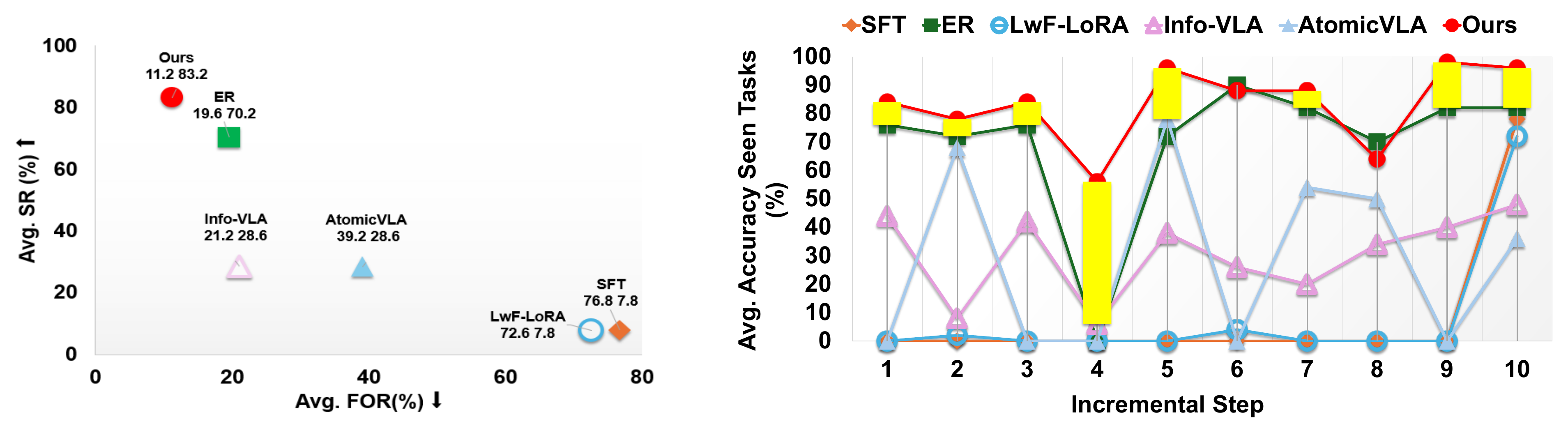}
  \caption{\textbf{Process-level analysis.} Left: stability--plasticity trade-off. Right: incremental accuracy over tasks. 
  }
  \label{fig:process}
\end{figure*}

\begin{table*}[t]
\centering
\caption{Comparison of success rate (SR, \%) under the LifelongVLA setting. Joint is an upper bound and is excluded from ranking. Red/blue denote the best/second-best lifelong methods.}
\label{tab:lifelongvla_sr}
\footnotesize
\setlength{\tabcolsep}{6.0pt}
\renewcommand{\arraystretch}{0.92}
\begin{adjustbox}{max width=\textwidth}
\begin{tabular}{l|cccccccccc|c}
\toprule
Method & 1 & 2 & 3 & 4 & 5 & 6 & 7 & 8 & 9 & 10 & Avg. \\
\midrule
Joint & 100 & 92 & 42 & 92 & 100 & 100 & 88 & 82 & 100 & 100 & 89.6 \\
\midrule
SFT (sequence fine-tune all) & 0 & 0 & 0 & 0 & 0 & 0 & 0 & 0 & 0 & 78 & 7.8 \\
LwF-LoRA~\citep{li2017learning}(2017) & 0 & 2 & 0 & 0 & 0 & 4 & 0 & 0 & 0 & 72 & 7.8 \\
ER~\citep{chaudhry2019tiny}(2019) & \second{76} & \second{72} & \second{76} & 0 & 72 & \best{90} & \second{82} & \best{70} & \second{82} & \second{82} & \second{70.2} \\
Info-VLA~\citep{zhao2026information}(2026) & 44 & 8 & 22 & \second{6} & 38 & 26 & 20 & 34 & 40 & 48 & 28.6 \\
AtomicVLA~\citep{zhang2026atomicvla}(2026) & 0 & 68 & 0 & 0 & \second{78} & 0 & 54 & 50 & 0 & 36 & 28.6 \\
\midrule
\rowcolor{gray!15}
\textbf{Ours} & \best{84} & \best{78} & \best{84} & \best{56} & \best{96} & \second{88} & \best{88} & \second{64} & \best{98} & \best{96} & \best{83.2} \\
\bottomrule
\end{tabular}
\end{adjustbox}
\end{table*}

\begin{table*}[t]
\centering
\caption{Comparison of forgetting rate (FOR, \%) under the LifelongVLA setting. Lower is better; red/blue denote the lowest/second-lowest values.}
\label{tab:lifelongvla_for}
\footnotesize
\setlength{\tabcolsep}{7.0pt}
\renewcommand{\arraystretch}{0.92}
\begin{adjustbox}{max width=\textwidth}
\begin{tabular}{l|cccccccccc|c}
\toprule
Method & 1 & 2 & 3 & 4 & 5 & 6 & 7 & 8 & 9 & 10 & Avg. \\
\midrule
SFT(sequence fine-tune all) & 96 & 92 & 32 & 100 & 96 & 86 & 82 & 88 & 96 & 0 & 76.8 \\
LwF-LoRA~\citep{li2017learning}(2017) & 96 & 78 & \second{10} & 94 & 90 & 94 & 82 & 84 & 98 & 0 & 72.6 \\
ER~\citep{chaudhry2019tiny}(2019) & \second{22} & 22 & \second{10} & 62 & 22 & \best{4} & \second{10} & \second{26} & \second{18} & 0 & \second{19.6} \\
Info-VLA~\citep{zhao2026information}(2026) & 28 & 18 & \best{4} & \second{46} & 24 & 44 & 16 & \best{6} & 26 & 0 & 21.2 \\
AtomicVLA~\citep{zhang2026atomicvla}(2026) & 76 & \best{8} & 34 & 70 & \second{6} & 92 & \best{8} & 28 & 70 & 0 & 39.2 \\
\midrule
\rowcolor{gray!15}
\textbf{Ours} & \best{14} & \second{10} & \best{4} & \best{36} & \best{2} & \second{8} & 12 & 28 & \best{0} & 0 & \best{11.4} \\
\bottomrule
\end{tabular}
\end{adjustbox}
\end{table*}

\begin{figure}[t]
  \centering
  \includegraphics[width=1\linewidth]{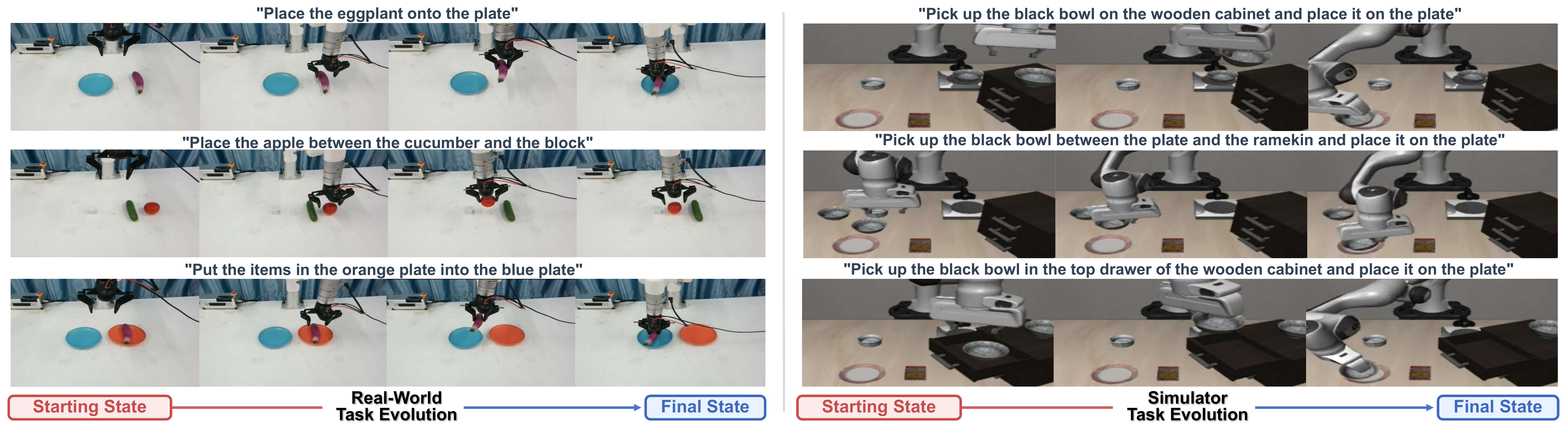}
  \caption{\textbf{Real-world and simulator task evolution.} The left panel shows real-world rollouts and the right panel shows simulator rollouts. Each row illustrates language-conditioned start-to-final-state execution.}
  \label{fig:real_sim}
\end{figure}

\begin{table}[t]
\centering
\caption{Complexity and ablation analysis under the LifelongVLA setting. The first three rows isolate the effects of latent replay representation and dual LoRA design in \method{}, while the remaining rows compare representative baselines in terms of memory cost, trainable parameters, average SR( \%), and average FOR( \%).}
\label{tab:complexity_ablation}
\footnotesize
\setlength{\tabcolsep}{5.5pt}
\renewcommand{\arraystretch}{0.92}
\begin{adjustbox}{max width=0.98\linewidth}
\begin{tabular}{lcccc}
\toprule
Method / Variant & Memory Cost & Trainable Params & Avg. SR & Avg. FOR \\
\midrule
Ours(raw data replay) & 167.62 MiB / task & 49.99M & 81.8 & 18.44 \\
Ours(single LoRA) & 95.70 MiB / task & 49.99M & 74.6 & 21.78 \\
\rowcolor{gray!15}
\textbf{Ours(latent replay + dual LoRA)} & \textbf{95.70 MiB / task} & \textbf{99.97M} & \textbf{83.2} & \textbf{11.4} \\
\midrule
LwF-LoRA~\citep{li2017learning}(2017) & 99.98 MiB / task & 49.99M & 7.8 & 72.6 \\
AtomicVLA~\citep{zhang2026atomicvla}(2026) & 452.98 MiB / task & 226.49M & 28.6 & 39.2 \\
Info-VLA~\citep{zhao2026information}(2026) & 138.32 MiB / task & 49.99M & 28.6 & 21.2 \\
\bottomrule
\end{tabular}
\end{adjustbox}
\end{table}

\section{Experiments}
\label{sec:experiments}

\paragraph{Experimental settings and metrics.}
We evaluate \method{} on 10 tasks selected from the LIBERO benchmark~\citep{liu2023libero} for incremental skill learning under the LifelongVLA setting. Each incremental step introduces one new task. For each task, the policy is trained only on the current task data with access to a bounded replay cache from previous tasks, and is evaluated on all seen tasks; joint training with all task data is reported as an upper bound. As shown in \cref{fig:method_overview}, we use a frozen PaliGemma backbone with Gemma 2B and a Gemma 300M continuous action decoder for 7D robot action ~\citep{pi2024pi0}. All lifelong methods are trained for 10,000 steps per task, whereas joint training is run for 40,000 steps until convergence. We set $r=16$, $\gamma=16$, $M=500$, $c_t=50$, $\lambda=1.0$, and $\beta=0.1$. For fair comparison, all methods use the same task order, action space, and simulation protocol, with task identity unavailable during testing. We report the final average success rate as $\mathrm{SR}=\frac{1}{T}\sum_{i=1}^{T}a_{T,i}$, where $a_{t,i}$ denotes the success rate on task $i$ after training through task $t$. Forgetting is measured as $\mathrm{FOR}_{i}=\max_{l\in\{i,\ldots,T-1\}}a_{l,i}-a_{T,i}$ for $i<T$, and the average forgetting rate over old tasks is $\mathrm{FOR}=\frac{1}{T-1}\sum_{i=1}^{T-1}\mathrm{FOR}_{i}$~\citep{chaudhry2018riemannian}. We also report process-level incremental accuracy, $\mathrm{ACC}_{t}=\frac{1}{t}\sum_{i=1}^{t}a_{t,i}$.

\subsection{Main Comparison and Process-Level Analysis}
We compare \method{} with representative lifelong learning and VLA adaptation baselines, including sequential fine-tuning (SFT), LwF-LoRA~\citep{li2017learning} with output distillation, ER~\citep{chaudhry2019tiny} with episodic replay, Info-VLA~\citep{zhao2026information} with information-guided regularization, and AtomicVLA\citep{zhang2026atomicvla} with skill-level expert specialization. As shown in \cref{tab:lifelongvla_sr,tab:lifelongvla_for}, \method{} achieves the best average SR of 83.2\% and the lowest average FOR of 11.4\%, improving over the strongest baseline ER by 13.0 points in SR and 8.2 points in FOR. SFT and LwF-LoRA obtain only 7.8\% average SR and suffer severe early-task forgetting, indicating that direct sequential fine-tuning or distillation alone is insufficient. Info-VLA and AtomicVLA improve some individual skills, but their average SR remains 28.6\%, suggesting that information-guided regularization or skill-level expert specialization alone does not fully resolve the stability--plasticity trade-off. At the task level, \method{} achieves strong performance on most skills, including 96\% on task 5, 98\% on task 9, and 96\% on task 10; task 9 further obtains a FOR of $0$, indicating almost no forgetting after sequential learning. \Cref{fig:process} provides a process-level view, where \method{} is closest to the upper-left ideal region in the stability--plasticity plot and maintains a stable incremental accuracy curve as more tasks arrive. These results show that dual-timescale adaptation and stochastic prefix replay jointly support efficient skill expansion while mitigating forgetting throughout continual learning.

\subsection{Complexity and Ablation}
\Cref{tab:complexity_ablation} reports memory usage, trainable parameters, and ablation results. We ablate the two key modules of \method{}: \textbf{Cache-Efficient Stochastic Replay} by comparing raw data replay, which stores full trajectories, with latent replay, and \textbf{Dual-Timescale LoRA Gating} by replacing dual LoRA with a single LoRA adapter.Latent replay reduces memory from 167.62 MiB/task to 95.70 MiB/task with the same 49.99M trainable parameters, while maintaining comparable SR and lower FOR, showing that compact cached prefixes with stochastic suffix recomputation can retain old skills without full trajectories. Dual LoRA further improves SR and reduces FOR over single LoRA, confirming the benefit of separating short-term plastic adaptation from long-term stable consolidation. Although dual LoRA increases trainable parameters to 99.97M, \method{} remains substantially more compact than AtomicVLA, which uses 226.49M trainable parameters and 452.98 MiB/task memory. Compared with LwF-LoRA and Info-VLA, \method{} uses comparable or lower memory while achieving much higher SR and lower FOR, demonstrating a better balance between efficiency and continual-learning performance.

\subsection{Real-Robot Evaluation}
We further evaluate \method{} on a five-task real-robot stream with an xArm robot, where each task is introduced sequentially. For each task, we collect 50 training episodes and evaluate the policy with 50 evaluation episodes per task on all seen tasks after each new task is learned. The policy learns language-conditioned manipulation from visual observations and is tested without task identity. After learning all five tasks, \method{} achieves success rates above 80\% on every task, indicating that the proposed retention mechanism remains effective beyond simulated benchmarks. As shown in \cref{fig:real_sim}, the qualitative rollouts further show successful start-to-final-state executions in real-world and simulator environments. Detailed task-wise success rates, forgetting rates, and deployment videos are provided in the supplementary material.

\section{Conclusion}
\label{sec:conclusion}
We presented \method{}, a lifelong VLA method for sequential robotic manipulation in open-world settings. To balance new-skill plasticity and old-skill stability, \method{} introduces \textbf{Dual-Timescale LoRA Gating}, which composes short- and long-term LoRA pathways with a task-aware weight-level gate. To reduce replay overhead, \method{} further develops \textbf{Cache-Efficient Stochastic Replay}, which stores compact latent prefixes and reconstructs replay suffixes under fresh diffusion variables. Experiments on lifelong manipulation tasks show that \method{} achieves higher final success rates, lower forgetting, and efficient memory usage with moderate trainable parameters compared with existing baselines. 

\section{Limitations and Future Work}

While our method demonstrates effective lifelong adaptation across sequential manipulation tasks, the current evaluation remains limited in task scale and diversity. Real-world lifelong learning may involve longer task streams, more diverse object categories, richer scene layouts, and non-uniform data collected over extended deployment. Future work could evaluate longer and more heterogeneous task sequences, randomize task orders, and report mean and variance across multiple streams to better assess long-term stability and robustness.

Although our experiments use clear language instructions, real users may provide more diverse, ambiguous, or conversational commands. Different expressions can share the same intent, while some may involve ellipsis, coreference, or contextual dependencies. Future work could introduce paraphrased, conversational, and context-dependent instructions to evaluate language robustness. Language augmentation or instruction rewriting may further improve the model's ability to handle diverse user expressions in lifelong VLA learning.

\clearpage
\bibliographystyle{plainnat}
\bibliography{references}
\end{document}